\documentclass[manuscript,screen]{acmart}
\pdfoutput=1
\AtBeginDocument{%
  \providecommand\BibTeX{{%
    \normalfont B\kern-0.5em{\scshape i\kern-0.25em b}\kern-0.8em\TeX}}}

\setcopyright{acmlicensed}
\copyrightyear{2025}
\acmYear{2025}
\acmDOI{XXXXXXX.XXXXXXX}

\usepackage{graphicx}
\usepackage{tabularx}
\usepackage{adjustbox}
\usepackage{longtable} 
\usepackage{rotating}
\usepackage{multirow}
\usepackage{float}
\usepackage{array}
\usepackage{booktabs}
\usepackage[normalem]{ulem}
\useunder{\uline}{\ul}{}
\usepackage{longtable}

\title[A Comprehensive Survey on Legal Summarization]{A Comprehensive Survey on Legal Summarization: Challenges and Future Directions}

\author{Mousumi Akter}
\affiliation{%
  \institution{TU Dortmund University \& Research Center Trustworthy Data Science and Security}
  \country{Germany}
  }
\email{mousumi.akter@tu-dortmund.de}

  \author{Erion {\c C}ano}
\affiliation{%
  \institution{Ruhr University Bochum \& Research Center Trustworthy Data Science and Security}
  \country{Germany}
  }
\email{erion.cano@ruhr-uni-bochum.de}

  \author{Erik Weber}
\affiliation{%
  \institution{TU Dortmund University \& Research Center Trustworthy Data Science and Security}
  \country{Germany}
  }
\email{erik.weber@tu-dortmund.de}

  \author{Dennis Dobler}
\affiliation{%
  \institution{TU Dortmund University \& Research Center Trustworthy Data Science and Security}
  \country{Germany}
  }
\email{dobler@statistik.tu-dortmund.de}

  \author{Ivan Habernal}
\affiliation{%
  \institution{Ruhr University Bochum \& Research Center Trustworthy Data Science and Security}
  \country{Germany}
  }
\email{ivan.habernal@ruhr-uni-bochum.de}

\begin{document}

\begin{abstract}
This article provides a systematic up-to-date survey of automatic summarization techniques, datasets, models, and evaluation methods in the legal domain. Through specific source selection criteria, we thoroughly review over 120 papers spanning the modern `transformer' era of natural language processing (NLP), thus filling a gap in existing systematic surveys on the matter. We present existing research along several axes and discuss trends, challenges, and opportunities for future research.
\end{abstract}



\maketitle

\section{Introduction}
\begin{figure*}[t]
    \centering
    \includegraphics[width=0.9\linewidth]{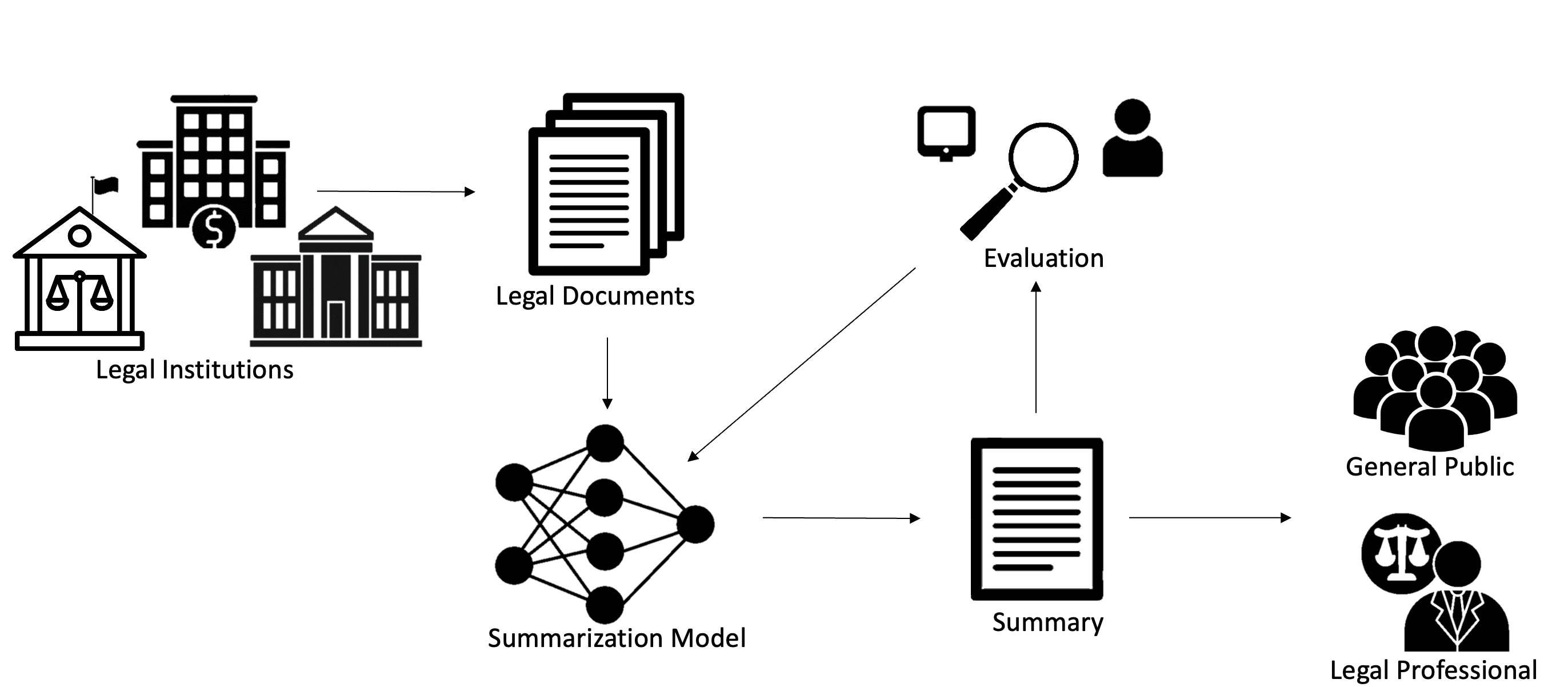}
    \caption{Schematic Legal Summarization Pipeline: Legal summarization pipelines process lengthy, structurally diverse documents to generate summaries, serving both general audiences and domain experts.}
    \vspace{-5 mm}
    \label{fig:pipeline}
\end{figure*}

In the legal domain, language is not just a means of communication, but rather the currency of the profession. The constant engagement with extensive written materials is fundamental and immensely time-consuming \cite{ThomsonReuters2023}. Legal professionals often spend hours, if not days, combing through documents to find precedents or relevant cases that could be pivotal to their current cases. This laborious process is a significant part of the workload of legal professionals like lawyers and judges, taking up lots of time that could be invested otherwise. Automatic summarization tools could help to condense lengthy legal documents into concise summaries, helping to save both time and costs. Moreover, integrating advanced Natural Language Processing (NLP) techniques into legal research holds significant promise for democratizing access to legal information. 
Figure~\ref{fig:pipeline} shows the general pipeline for legal summarization.

Compared to other domains, legal texts present unique challenges that distinguish them from other document types. Legal documents tend to be longer and more detailed than those from other domains. They feature complex language with abundant domain-specific terminology, abbreviations, and extensive use of citations or references to related documents, leading to this language often being referred to as 'legalese'. Moreover, legal documents exhibit diverse structural formats based on their country of origin. For example, the legal documentation from the United States differs significantly in structure from that of India. This variation can pose considerable challenges when developing a generalized tool for summarizing legal documents effectively. 

Legal summarization can generally be divided into three main aspects: region-specific legal documents, summarization strategies and summarization methods. Region-specific legal summarization focuses on developing tools that can handle documents from particular countries or languages. Additionally, general summarization strategies have also been identified in legal summarization, including extractive, abstractive, and hybrid approaches. The extractive approach involves directly copying significant sentences from the source document and combining them to create the output~\cite{DBLP:conf/acl/0001L16}. In contrast, the abstractive approach mimics human understanding by interpreting the source document and producing a summary based on its key concepts~\cite{DBLP:conf/emnlp/RushCW15}. The hybrid approach aims to combine the strengths of both methods by rewriting a summary that utilizes selected important content extracted from the source document. On the other hand, summarization methods vary from rank-based, graph-based, transformer-based, and others.

Despite significant progress in the research of legal document summarization \cite{DBLP:journals/air/KanapalaPP19, DBLP:conf/ecir/BhattacharyaHRP19, agrawal2020legal, DBLP:journals/csr/JainBB21, DBLP:conf/ijcnlp/ShuklaBPMGGG22, ariai2024natural}, the field lacks a comprehensive and timely survey.
We address this gap in our article as follows: 
\begin{description}
\item[Comprehensive overview of summarization research] We provide a comprehensive survey on the current state of summarization approaches specifically designed for the legal domain. This encompasses three main areas of legal summarization: existing methods and approaches, available datasets, and evaluation metrics used. 
\item[Comprehensive analysis of research trends in legal summarization] We conduct a thorough examination of research trends in legal summarization, exploring three aspects such as region-specific trends, legal summarization strategies, and various methodologies. This analysis aims to provide valuable insights into the evolving landscape of legal summarization practices.
\item[Challenges and future directions] We examine the limitations present in current methodologies and propose future research directions that hold significant promise. Key areas for exploration include enhancements in model design, improvements in the quality and diversity of datasets, as well as the feasibility of automated evaluation metrics alongside human evaluation strategies.
\end{description}

This survey is organized as follows. First, an overview of the paper selection methodology for the survey is presented in Section~\ref{survey_method}. Next, Section~\ref{datasets} provides a detailed overview of the available legal summarization datasets. The later part of the survey offers a comprehensive review of research trends in legal summarization from three perspectives: regional specifics, discussed in Section~\ref{region}; approaches for legal summarization strategies, covered in Section~\ref{strategy}; and legal summarization methodologies, detailed in Section~\ref{method}. In Section~\ref{metric}, we examine the trends in evaluation approaches for legal summarization and discuss the available metrics. Section~\ref{future} addresses the limitations of current approaches and outlines promising directions for future research in this field. Finally, Section~\ref{conclude} concludes the survey.

\section{Paper selection methodology}~\label{survey_method}
The steps we followed during the paper selection phase are similar to the ones suggested in the guidelines for systematic literature reviews defined by Kitchenham and Charters \cite{kitchenham2007guidelines}, commonly followed in similar studies \cite{ccano2017hybrid,DBLP:journals/concurrency/FigueroaVRM15}. The purpose of using a popular and reproducible methodological approach is to ensure that the treatment of the research topic is verifiable and unbiased.  

 
We chose eight main digital repositories of primary studies and four more secondary (indexing) digital repositories, shown in Table~\ref{tab:primlibraries}.

\begin{table}
\begin{tabular}{l l}
\toprule
\textbf{Primary repositories} & \\
ACL & \url{https://aclanthology.org} \\
TACL & \url{https://direct.mit.edu/tacl} \\  
Springer Link & \url{https://link.springer.com} \\
ScienceDirect & \url{https://www.sciencedirect.com} \\
IEEExplore & \url{http://ieeexplore.ieee.org} \\
ACM Digital Library & \url{http://dl.acm.org} \\
JURIX 2023 & \url{https://jurix23.maastrichtlawtech.eu} \\
Artificial Intelligence and Law & \url{https://link.springer.com/journal/10506} \\
\midrule
\textbf{Secondary repositories} & \\
Scopus & \url{http://www.scopus.com} \\
Semantic Scholar & \url{https://www.semanticscholar.org} \\  
Google Scholar & \url{https://scholar.google.com} \\
DBLP & \url{https://dblp.org} \\
\bottomrule
\end{tabular}
\caption{Main libraries searched for primary studies.\label{tab:primlibraries}}
\end{table}

\begin{table}
\begin{tabular}{l l}
\hline 
\bf Keyword & \bf Synonyms \\
\hline
legal & law, legislative, jurisdiction \\
text & textual, text-based, document, contract, decisions, ruling \\  
summarization & summary \\
dialogue & question and answering \\
\hline
\end{tabular}
\caption{Set of keywords and synonyms for search.\label{tab:searchterms}}
\end{table}

Furthermore, we defined (``legal'', ``text'', ``summarization'', ``dialogue'') as the set of the main keywords. Each of them was augmented with synonyms. The full list of search terms is shown in Table~\ref{tab:searchterms}. The search query derived from the keywords and synonyms is
(``legal'' OR ``law'' OR ``legislative'') AND (``summarization'' OR ``summary'') AND (``of'' OR ``about'') AND (``dialogues'' OR ``question and answering'' OR ``text'' OR ``textual'' OR ``text-based'' OR ``document'').

The search query was used in the search engines of the digital libraries; different numbers of papers were returned by each of them. The retrieval process was conducted during July 2024. 

To be objective in deciding which papers to include and which not, we defined a set of inclusion and exclusion criteria presented in Table~\ref{tab:incexc}. The inclusion and exclusion criteria reflect the objectives or the survey. They helped us to thematically focus on the most recent and relevant studies, to see at what scale Transformer-based models have been adopted in the legal domain, and to find sources of datasets and evaluation metrics. 

\begin{table}[ht]
\begin{tabular}{l}
\hline 
\bf Inclusion criteria \\
\hline
Papers on methods, datasets and metrics for legal summarization \\
Papers published in general conferences/journals and legal/summarization/evaluation workshops  \\ 
Papers published from 2017 to 2024 (Transformer era) \\ 
Papers with datasets and metrics, we include paper before 2017 \\ 
\hline 
\bf Exclusion criteria \\
\hline
Papers not about summarization \\
Papers outside the legal domain \\
Gray literature (not published in any reputable venue) \\
Papers published before 2017 \\
Publications not in English \\
\hline
\end{tabular}
\caption{Inclusion and exclusion criteria.\label{tab:incexc}}
\end{table}

Initially, we started with a coarse selection of the papers. During this phase, we checked titles and abstracts only. Making use of the inclusion and exclusion criteria and checking titles and abstracts, we reached to a collection of 262 papers. Duplicates were removed and a more detailed inspection of each paper was conducted, carefully reading the abstract and other sections. Besides relevance, completeness in terms of task definition, description of the proposed model or method, and presentation of results were considered. The inclusion and exclusion decisions were made in agreement between the first three authors. At the end of this phase, a total of 123 final papers were reached. Some publication details of those papers are presented in the Appendix.

\section{Datasets}~\label{datasets}
\begin{table*}[!htb]
\scriptsize
\centering 
\begin{tabularx}{\textwidth}{m{2cm} m{2.5cm} m{1.5cm} X m{1.5cm} m{1.5cm} }
\hline
\textbf{Dataset} & \textbf{Source} & \textbf{Domain} & \textbf{Short description} & \textbf{Language(s)} & \textbf{Size (\# Docs)} \\ \hline
HOLJ Corpus \newline \citet{grover-etal-2004-holj} & n.a. anymore & Court rulings & Judgements from the British House of Lords with human written summaries & English & 188 (153) \\ \hline
BrazilianBR \newline \citet{feijo2018rulingbr} & \url{https://github.com/diego-feijo/rulingbr} & Court rulings & Rulings from the Brazilian Supreme Court. Every document consists of four parts, the Ementa can be used as abstractive reference summary & Portuguese & 10,623 \\ \hline
IN-Abs \newline \citet{DBLP:conf/ijcnlp/ShuklaBPMGGG22} & \url{https://github.com/Law-AI/summarization} & Court rulings & Judgements from the Indian Supreme Court with headnotes as abstractive reference summaries & English & 7,130 \\ \hline
IN-Ext \newline \citet{DBLP:conf/ijcnlp/ShuklaBPMGGG22} & \url{https://github.com/Law-AI/summarization} & Court rulings & Judgements from the Indian Supreme Court with human-written extractive reference summaries. Summaries cover 7 rhetorical segments. & English & 50 \\ \hline
UK-Abs \newline \citet{DBLP:conf/ijcnlp/ShuklaBPMGGG22} & \url{https://github.com/Law-AI/summarization} & Court rulings & Judgements from the UK Supreme Court with press summaries as abstractive reference summaries & English & 793 \\ \hline
AustLII \newline \citet{legal_case_reports_239}  & \url{https://archive.ics.uci.edu/dataset/239/legal+case+reports} & Court rulings & Cases from the Federal Court of Australia with catchphrases, citations sentences, citation catchphrases, and citation classes. & English & 3,890\\ \hline
LegalSum (German) \newline \citet{glaser-etal-2021-summarization} & \url{https://github.com/sebimo/legalsum} & Court rulings & German court rulings with guiding principles as reference summaries. & German & 100,018 \\ \hline
Multi-LexSum \newline \citet{10.5555/3600270.3601226}  & \url{https://multilexsum.github.io/} & Court rulings & Documents for 4,534 civil right cases from the CRLC with abstractive summaries in three different lengths. & English & 40,119 (9,280) \\ \hline
CanLII & \url{https://www.canlii.org/en} & Court rulings & Cases from Canadian courts gathered by the Canadian Legal Information Institute. & English & 28,733 \\ \hline
LexAbSumm \newline \citet{t-y-s-s-etal-2024-lexabsumm} & \url{https://huggingface.co/datasets/MahmoudAly/LexAbSumm} & Court rulings & Aspect-Judgement-Summary triplets from the ECHR & English & 1,053 \\ \hline
ITA-CaseHold \newline \citet{10.1145/3594536.3595177} & \url{https://github.com/dlicari/ITA-CASEHOLD} & Court rulings & Pairs of judgments and holdings from the archives of Italian Administrative Justice & Italian & 1,101 \\ \hline
GreekLegalSum & \url{https://huggingface.co/datasets/DominusTea/GreekLegalSum} & Court rulings & Judgements and summaries from the Supreme Civil and Criminal Court of Greece & Greek &  8,395 cases + summaries, 6,370 with classification \\ \hline
CLSum-CA \newline \citet{liu2024low} & \url{https://github.com/StevenLau6/CLSum} & Court rulings & Judgements and case briefs from 2018 to 2023 from the Supreme Court of Canada & English & 192\\ \hline
CLSum-HK \newline \citet{liu2024low} & \url{https://github.com/StevenLau6/CLSum} & Court rulings & Cases from multilevel courts in Hong Kong including press summaries & English & 793\\ \hline
CLSum-AUS \newline \citet{liu2024low} & \url{https://github.com/StevenLau6/CLSum} & Court rulings & Australian judgment documents and their summaries from 2005 to 2023 & English & 1,019\\ \hline
MILDSum \newline \citet{datta-etal-2023-mildsum} & \url{https://github.com/Law-AI/MILDSum} & Court rulings & Case judgments from multiple High Courts and the Supreme Court of India & English, Hindi & 3,122\\ \hline
RechtspraakNL \newline \citet{schraagen-etal-2022-abstractive}     &  \url{https://git.science.uu.nl/n.vandeluijtgaarden/legal-text-summarization} & Court rulings & Large collection of Dutch cases & Dutch & ~400k \\ \hline
Privacy Policy Summarization Dataset \newline \citet{bannihatti-kumar-etal-2022-towards} & \url{https://github.com/awslabs/summarization-privacy-policies} & Privacy Policy & Sections of large English privacy policy dataset with short title summaries & English & 24,000\\ \hline
KorCase\_Summ & \url{https://github.com/saekomdalkom/KorCase_summ} & Privacy Policy & Precedents released on the Korean Court Comprehensive Legal Information website & Korean & 72,537 \\ \hline
Amicus Briefs \newline \citet{bajaj-etal-2021-long} & \url{https://www.publichealthlawcenter.org/litigation-tracker} & Legal Briefs & Arguments that
the court should consider for a case and their summary & English & 120 \\ \hline
BillSumm \newline \citet{kornilova-eidelman-2019-billsum} & \url{https://github.com/FiscalNote/BillSum} & Legislative bill & Mid-length bills with 5,000 to 20,000 characters length from the US congress or the state of California & English & 22,218 + 1,237 \\ \hline
JRC Acquis & \url{https://mediatum.ub.tum.de/1446654} & Legislative acts & Legislative documents of the European Parliament with short 1-3 sentences summaries & CS, DE, EN, ES, FR, IT, SV & $\leq 22,751$ docs per language \\ \hline
EUR-Lex-Sum \newline \citet{aumiller2022eurlexsum} & \url{https://github.com/achouhan93/eur-lex-sum} & Legislative acts & Legal regulatory acts passed by the European Union with abstractive reference summaries & 24 official EU languages & $\leq$ 1,505 docs per language \\ \hline
EUR-LexSum \newline \citet{DBLP:conf/sigir/KlausHNABH22} & \href{https://github.com/svea-klaus/Legal-Document-Summarization}{https://github.com/svea-klaus/Legal-Document-Summarization} & Legislative acts & Legal regulatory acts passed by the European Union & English & 4,595 \\ \hline
TL;DR/TOS;DR \newline \citet{manor-li-2019-plain}   &  \url{https://github.com/lauramanor/legal\_summarization} & Contracts & Software licenses and agreement texts with abstractive plain language summaries & English & 506 \\ \hline
\end{tabularx}
\caption{Overview of legal summarization datasets}
\label{tab:dataset}
\end{table*}
\newcolumntype{C}{>{\centering\arraybackslash}X}

While there is an extensive amount of unlabeled legal data, there are only a few datasets available for the task of legal summarization. These include a diverse range of documents such as contracts, legislative bills, and judicial decisions, sourced from various countries, institutions, and companies. This section is dedicated to offering a systematic overview of these documents, organizing them by type.

\subsection{Court rulings}

Court rulings are formal decisions issued by judges or judicial bodies in legal disputes, often serving as precedents or interpretations of law. They represent the largest group of documents within the domain of legal summarization.

\begin{itemize}

\item \citet{grover-etal-2004-holj} gathered a corpus of 188 judgments of the British House of Lords, out of which 153 judgments had the corresponding hand-written summaries. The dataset does not seem to be available anymore.

\item Various methods utilized the \textit{BrazilianBR} dataset proposed by \citet{feijo2018rulingbr}. It contains roughly 10,000 rulings from the Brazilian Federal
Supreme Court. Each document consists of four parts. The first part, the \textit{Ementa} is a brief summary of the case and the final decision, which can be used as the reference summary. 

\item \citet{DBLP:conf/ijcnlp/ShuklaBPMGGG22} proposed three novel datasets for legal summarization, two datasets for abstractive summarization and one dataset for extractive summarization. The \textit{IN-Abs} dataset contains 7,130 cases from the Indian Supreme Court collected from Legal Information Institute of
India\footnote{\url{http://www.liiofindia.org/in/cases/cen/INSC/}} with respective headnotes as reference summaries. Contrarily, the \textit{IN-Ext} dataset consists of 50 judgments from the Indian Supreme Court with human-generated extractive summaries through three legal professionals from the Gandhi School of Intellectual Property Law. These summaries are about one third of the length of the original document and were written with respect to seven rhetorical segments, which were labeled beforehand. Finally, the \textit{UK-Abs} dataset consists of 793 UK Supreme Court judgments\footnote{\url{https://www.supremecourt.uk/decided-cases/}} with their official press summaries, which can be utilized as the reference abstractive summaries of the documents. 

\item The Canadian Legal Information Institute (CanLII)\footnote{\url{https://www.canlii.org/en/}}is an organization that provides open online access to legislative decisions and documents, including summaries for many cases. The data was used in several publications such as \citet{elaraby2022arglegalsumm, villata2020using, elaraby-etal-2023-towards}

\item \citet{10.1145/3322640.3326728} utilized cases from the US Board of Veterans' Appeals (BVA)\footnote{\url{https://www.bva.va.gov/}}, particularly those dealing with service-connected post-traumatic stress disorder. For a number of these cases, extractive summaries were manually created by law experts. The data used for the paper were publicly accessible\footnote{\url{https://github.com/luimagroup/bva-summarization/}}.

\item \citet{galgani2010} released their dataset with about 4000 cases of the Federal Court of Australia. The dataset includes catchphrases for every document, where the catchphrases can be used as reference summaries.

\item \citet{Shen2022MultiLexSum} proposed the Multi-LexSum dataset, which contains 40,000 source documents for roughly 4,500 federal U.S. civil rights lawsuits. It contains muti-document summaries at three different levels of granularity, i.e., long summaries (for every case), small summaries (for about 70\% of the cases) and tiny summaries (for about 36\% of the cases). The data stems from the Civil Rights Litigation Clearinghouse (CRIC)\footnote{\url{https://clearinghouse.net/}}, an NGO dedicated to provide access to resources and information relating to civil rights litigation.

\item \citet{glaser-etal-2021-summarization} proposed a large-scale dataset of roughly 100,000 German court rulings with short summaries. The data was crawled from several sources\footnote{\url{https://www.otto-schmidt.de/}, \url{gesetze-bayern.de}, \url{justiz.nrw.de}} and contains cases of different German courts and legal areas.

\item \citet{info14040250} published a dataset of 8,395 cases from the Greek Supreme Court. 6,370 of these cases are classified with one or more tags. Every case consists of a heading, a summary written by a legal expert, an introduction to the case, a legal analysis as well as the final decision of the judges. 

\item \citet{liu2024low} released the CLSum dataset, which consists of four different sub-datasets for legal summarization in low-resource settings. CLSum-CA contains 192 cases collected from the Supreme Court of Canada between 2018 and 2023. Case briefs are used as reference summaries. CLSum-HK contains 233 judgments and their respective press summaries from the legal reference system of Hong Kong between 2012 and 2023. CLSum-AUS covers 1,019 cases from the High Court of Australia between 2005 and 2023, collected from the courts website. The fourth subset, CLSum-UK stems from \citet{DBLP:conf/ijcnlp/ShuklaBPMGGG22} and contains 793 judgments from the United Kingdom Supreme Court.

\end{itemize}

\subsection{Legislative documents}

\begin{itemize}

\item \citet{kornilova-eidelman-2019-billsum} proposed the \textit{BillSum} dataset of US Congressional and California state bills. Upon release it consisted of 22,218 bills of the US Congress as well as 1,237 California state bills, both with human-written reference summaries. The data was sourced from the Govinfo service, which is facilitated by the United States Government Publishing Office (GPO)\footnote{\url{https://github.com/unitedstates/congress?tab=readme-ov-file}}. The dataset is publicly
available\footnote{\url{https://github.com/FiscalNote/BillSum}}.

\item The \textit{EUR-Lex-Sum} dataset proposed by \citet{aumiller2022eurlexsum} is a multilingual dataset that contains legal acts issued by the European Union and their respective human-written summaries for each of the 24 official languages within the EU. Depending on the language, between 391 (Irish) and 1,505 (French) document-summary pairs are available. The data and the code are available online.\footnote{\url{https://github.com/achouhan93/eur-lex-sum}}

\item \citet{10.1145/3477495.3531872} published a similar resource, also based on EUR-Lex documents. In contrast, the \textit{EUR-LexSum} dataset is a monolingual corpus that consists of 4,595 documents with their respective summaries.

\item \citet{elnaggar2018multitaskdeeplearninglegal} released a corpus derived of the original large-scale JRC Acquis corpus by the Joint Research Centre of the European Commission, which contains legislative documents of the European Parliament since 1958. The data comes in 7 languages (cs, de, en, es, fr, it, sv), with up to 22,000 documents per language.

\end{itemize}

\subsection{Privacy policies}

\begin{itemize}
\item \citet{keymanesh2020toward} published a dataset that contains the privacy policies of 151 companies, each sentence annotated with a label reflecting its privacy-related 'risk class'\footnote{\url{https://github.com/senjed/Summarization-of-Privacy-Policies}}. These labels, as well as their reference plain language summaries, have been taken from the \textit{TOS;DR} website\footnote{\url{https://tosdr.org/}}, a site that evaluates and explains companies' privacy policies in plain English. 

\item Another dataset for privacy policy summarization was introduced by \citet{bannihatti-kumar-etal-2022-towards}. They leveraged a large English language privacy policy dataset originally published by \citet{Amos_2021} and sampled 20,000 policies from this dataset. Afterwards, 24,000 sections were randomly sampled and body-title pairs were extracted.
\end{itemize}

\subsection{Other domains}

In addition to court rulings, legislative documents, and privacy policies, legal summarization can also be applied to other areas such as legal opinions and legal briefs.

\begin{itemize}

\item \citet{bajaj-etal-2021-long} collected a dataset of 120 so-called Amicus Briefs, which include detailed case arguments that
the court should consider as source texts and their summaries as targets.
\end{itemize}

In addition to the discussion above, Table~\ref{tab:dataset} provides an overview of all datasets available for legal summarization.

\section{Region-specific legal summarization}\label{region}

\begin{figure}[!b]
    \centering
    \includegraphics[width=0.8\linewidth]{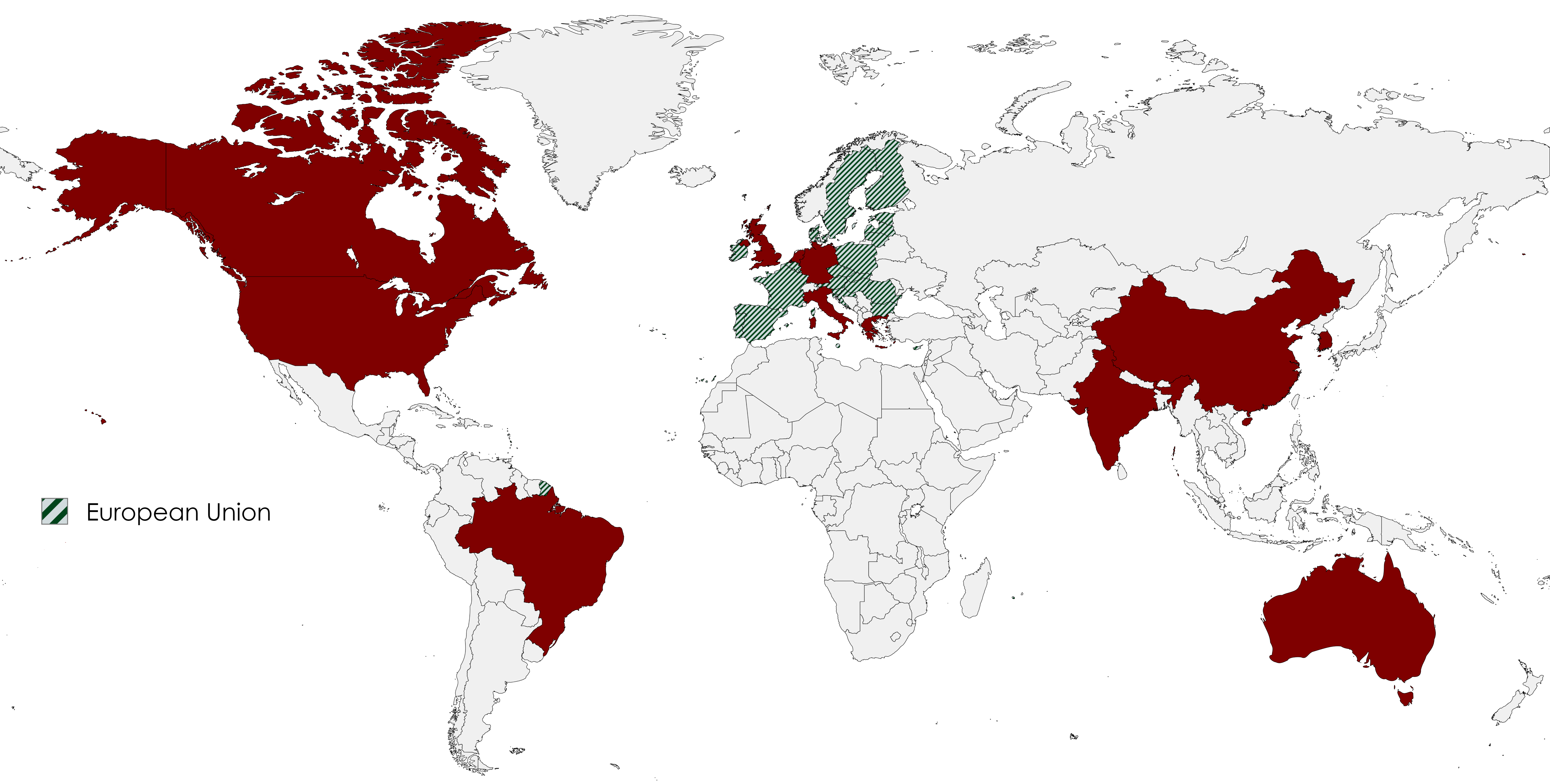}
    \caption{Countries identified in the collection of papers during the survey study}
    \label{fig:enter-label}
\end{figure}

The legal domain is highly region-specific and is shaped by unique cultural, linguistic, and legal traditions. Legal summarization research is no exception to this. The effectiveness of the summarization approaches depends heavily on their ability to handle regional nuances. Models designed for one legal system often cannot be seamlessly transferred to another, as legal texts are closely tied to the jurisdictions that produce them. For instance, Common Law jurisdictions like the United States or the United Kingdom heavily rely on precedents, resulting in lengthy and narrative case law. In contrast, civil law systems, as in Germany or Brazil, focus on statutes that demand concise and precise legal language. A summarization model trained on documents from a Common Law jurisdiction might fail when applied to the succinct and structured statutes of civil law countries.

One example of applying legal summarization techniques across multiple jurisdictions is \cite{DBLP:journals/csr/JainBB21} where they conducted a case study analyzing the challenges. The study highlighted significant difficulties in generalizing methods trained on legal documents from one region to another. 
Similarly, \cite{DBLP:conf/ijcnlp/ShuklaBPMGGG22} conducted research on legal document summarization for Indian and UK Supreme Courts. Their work demonstrates the need for jurisdiction-specific datasets, as models trained on one legal system often fail to generalize to another. Fine-tuning models like Legal-Pegasus on domain-specific corpora significantly enhances summarization performance, emphasizing the limitations of general-purpose models. 
Not only do legal summarization models face challenges when applied across different countries, but even within a single country, different courts pose distinct obstacles. This has been demonstrated with two Italian courts in \cite{achena2023legal}. It is also worthy mentioning that recent publications in the field of legal summarization like \cite{bajaj-etal-2021-long} have shifted the focus on low-resource languages and jurisdictions. A list of works and resources specific to certain regions of the world are provided below. 

\paragraph{English and international}
English is the dominant language when it comes to approaches for summarizing court rulings. Summarization datasets from Common Law countries (e.g., UK, USA, Canada, Australia, India) such as BillSum \cite{kornilova-eidelman-2019-billsum}, the HOLJ corpus \cite{grover-etal-2004-holj}, the CanLII database,\footnote{\url{https://www.canlii.org/en}} or the AustLII corpus \cite{legal_case_reports_239} have been utilized in various papers to train legal summarization models. For instance, the BillSum dataset was used in 14 papers we collected. Furthermore, legal documents from India, particularly judgments from the Indian Supreme Court represent another important source of data in legal summarization research. Despite India's multilingual context, these rulings are predominantly written in English. Datasets like IN-ABS and IN-EXT,\footnote{\url{https://github.com/Law-AI/summarization?tab=readme-ov-file}} which provide abstractive and extractive summaries, respectively, have been used in various studies.   

\paragraph{European specifics} European legal summarization approaches have been focused around a couple of available datasets and reflect the highly regulative nature of European legislation. Some of them include legislative acts passed by the European Union. One such datasets is EUR-Lex-Sum released by \cite{aumiller-etal-2022-eur}. It is a multi-lingual and cross-lingual collection which covers 24 European languages. Another very similar dataset is EUR-LexSum released in \cite{10.1145/3477495.3531872}. It comprises 4,595 documents, again based on legislative acts passed by the European Union. Contrary to EUR-Lex-Sum, the documents in EUR-LexSum are in English only. Multi-LexSum released in \cite{shen2022multilexsumrealworldsummariescivil} is another dataset pertaining to the European landscape. It includes 9,280 summaries of experts, based on publications from the CRLC (Civil Rights Litigation Clearinghouse). Multi-LexSum is distinct from other collections in its multiple target summaries, each at a different granularity. Experimenting with short summaries is also important in some cases. JRC-Acquis\footnote{Available at \url{https://mediatum.ub.tum.de/1446654}} is a collection of legislative documents of the European Parliament with short (e.g., 1-3 sentences long) summaries that can be utilized for that purpose. Other datasets such as LegalSum \cite{glaser-etal-2021-summarization} and ITA-CaseHold \cite{10.1145/3594536.3595177} contain documents in European languages (German and Italian, respectively) and they reflect the legal nuances of the respective countries. 

\paragraph{Asia} There are several corpora from different countries in east Asia included in our study. China judgment Online\footnote{\url{https://wenshu.court.gov.cn}} is one of them, consisting of both criminal and civil cases. CAIL2020 \cite{10.1162/dint_a_00094} is another collection of 13,531 court rulings. Mandarin Chinese is also represented in the Court Debte Dataset \cite{ji-etal-2023-cdd} which is a large collection of 30,481 court cases, totaling 1,144,425 utterances exchanged between the plaintiff and the defendant. Finally, another corpus coming from this region of this world is KorCase\_Summ\footnote{\url{https://www.law.go.kr/LSW/lawPetitionForm.do?subMenuId=79&menuId=13}} which includes precedents of the Korean Court.

\paragraph{Latin America} A good representative of the Latin America countries is BrazillianBR, a collecion of decisions from the supreme court of Brazil, dating between 2011 and 2018 \cite{feijo2018rulingbr}. Each document is in Portuguese and contains four parts, one of which is an abstractive summary which can be used as reference.

\section{Strategies for legal summarization}\label{strategy}
\begin{figure}
    \centering
    \includegraphics[width=0.5\textwidth]{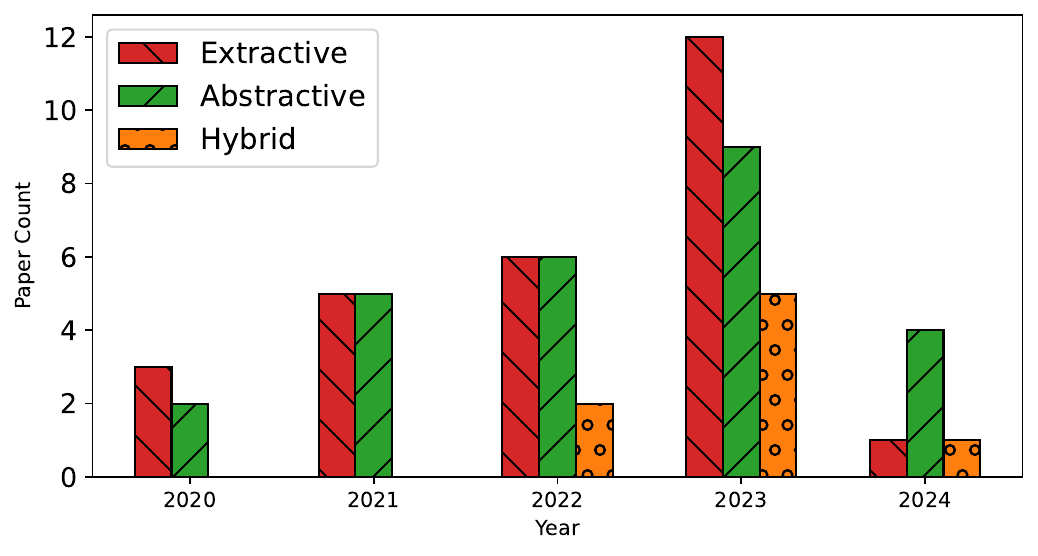} 
    \caption{Legal summarization research trends for last 5 years}
    \label{fig:trend}
\end{figure}

This section provides a thorough overview of legal summarization strategies, focusing on both extractive and abstractive methods. Additionally, we include a summary table in Appendix that encompasses all of the discussed methods for easy reference.

\subsection{Extractive legal summarization}

In legal summarization, extractive strategies have become increasingly popular. This approach involves extracting essential information from lengthy, domain-specific texts, such as judicial decisions, legal judgments, and contracts. Extractive methods are particularly effective in the legal field because they preserve the integrity of the original text, ensuring that critical legal content is not misinterpreted. We observed several highlights of approaches for legal summarization.

\paragraph{Rule-based approaches} Optimization techniques and heuristic algorithms play a significant role in ranking or selecting sentences for legal summarization, offering a rule-based and algorithmic approach with lower computational overhead compared to embedding-based methods. These techniques are particularly effective in tasks where domain knowledge can be encoded through explicit rules. For instance, researchers have utilized latent semantic analysis~\cite{DBLP:conf/icacci/MerchantP18, DBLP:conf/iccsci/SagumCCL23} and document-specific catchphrases~\cite{DBLP:conf/jurix/MandalBM021} to enhance extractive summaries. Bayesian optimization~\cite{DBLP:conf/fire/JainBB20} of TextRank hyperparameters has also been employed to refine the summarization process. Additionally, algorithm-based approaches like a reweighting mechanism for HipoRank~\cite{DBLP:conf/acl-nllp/ZhongL22} have gained attention, enabling dynamic updates to sentence importance scores based on the history of previously selected sentences, resulting in more accurate extractive summaries of legal documents. Furthermore,~\cite{DBLP:conf/cikm/DuanZYZLWWZS019} proposed an extractive summarization framework for court debates, utilizing an encoder-decoder neural network to jointly model debate utterances alongside additional information, such as legal knowledge graphs and semantic data, to enhance summarization performance. 

Further advancements incorporate hybrid and optimization-driven techniques. For instance, one study combines Legal-BERT sentence embeddings with Anonymous Walk Embeddings~\cite{DBLP:conf/icon-nlp/JainBB21} of the entire document graph through concatenation. This combined representation is then utilized in a multi-layer perceptron (MLP) to classify the summarization potential of sentences. Another method employs Integer Linear Programming (ILP)~\cite{10.1145/3462757.3466092} to optimize an objective function, generating summaries that capture the most informative sentences, ensure balanced representation across different thematic segments, and reduce redundancy. These developments illustrate the increasing complexity and sophistication of approaches aimed at overcoming the specific challenges associated with legal text summarization. Meanwhile,~\cite{DBLP:conf/iconip/NguyenNNLKNL21} begins by training an extractive summarization backbone model using standard supervised training. It then fine-tunes this backbone using reinforcement learning, incorporating a novel reward model that seamlessly integrates lexical, sentence, and keyword-level semantics into a single reward function.

\paragraph{Rank-based approaches} These approaches are popular for ranking essential legal information based on relevance or contextual scores~\cite{DBLP:journals/ail/JainBB24}. They often combine traditional information retrieval methods, such as BM25~\cite{Andrade2023BB25HLegalSumLB} and TF-IDF~\cite{polsley-etal-2016-casesummarizer}, with modern embedding models to enhance relevance. Classic methods are less computationally intensive and effectively complement neural models in processing long legal documents for extractive summarization.

\paragraph{Model-centric approaches}  These approaches that utilize pre-trained neural models, embeddings, or deep learning techniques often involve fine-tuning models based on architectures such as CNN~\cite{DBLP:journals/cacm/KrizhevskySH17}, LSTM~\cite{hochreiter1997long}, or Transformers~\cite{vaswani2017attention}. One method is query-based summarization~\cite{DBLP:conf/jurix/ZinNSSN23}, which extracts sentences relevant to predefined queries. After this process, the generated summary is provided to GPT-3.5 for information extraction.~\citet{purnima2023citation} used LegalBERT to obtain sentence embeddings for both citation sentences and judgment sentences. Then, cosine similarity is employed to select the summary sentence. They have various approaches for scoring the sentences. On the other hand,~\cite{DBLP:conf/lirai/BauerSGA23} used a Longformer encoder, pre-trained on the LegalBART objective using 6 million U.S. Court opinions.
Classic deep learning models such as CNN and LSTMs remain relevant in legal summarization. For instance, a method proposed in ~\cite{DBLP:conf/iconip/ChenYZ23} incorporates a sentence encoder, topic model, position encoder, TS-LSTM network, law article processing, and a sentence classifier for enhanced summarization. Moreover,~\cite{ANAND20222141} utilize weak supervision to label sentences as important or not, and then apply LSTM to generate the summary.

On the other hand,~\cite{10.1145/3322640.3326728} developed a CNN-based model that selects predictive sentences from legal case documents. The model classifies these sentences into types and employs maximal margin relevance alongside a summarization template to generate the summaries. 
Some studies also provide comparisons~\cite{rusiya2023implementation} of model-based approaches. For example, one study~\cite{DBLP:conf/sigir/KlausHNABH22} comparing BERT, DistilBERT, and RoBERTa against TextRank demonstrated that the former performed better, even when fine-tuned with limited data from the EUR-LexSum dataset.

\paragraph{Region-specific approaches} Region-specific adaptations often involve fine-tuning pre-trained models on datasets specific to a particular jurisdiction. In work~\cite{DBLP:conf/icccnt/TrivediTVJMD20}, the authors specifically focused on Indian legal documents and employs a Support Vector Machine (SVM) classifier to extract a summary of each legal document, which is then utilized to search for similar judgment cases and documents. Additionally, the study examines a Portuguese dataset~\cite{DBLP:conf/lacci/MedinaOFSROSSFF22}. The authors begin by preprocessing the documents and then perform extractive summarization of sentences using the PageRank algorithm. Following this, they employ a Bag-of-Words representation and Support Vector Classifier (SVC) to classify the documents into their respective categories. There has also been a specific focus on Italian extractive legal case summarization,~\cite{10.1145/3594536.3595177} fine-tuning the Italian-BERT, ItalianLEGAL-BERT, and Italian-LEGAL-BERT-SC models to predict the most relevant sentences and then create the summary.

\begin{figure}
    \centering
    \includegraphics[width=0.95\textwidth, trim=0 0 0 80, clip]{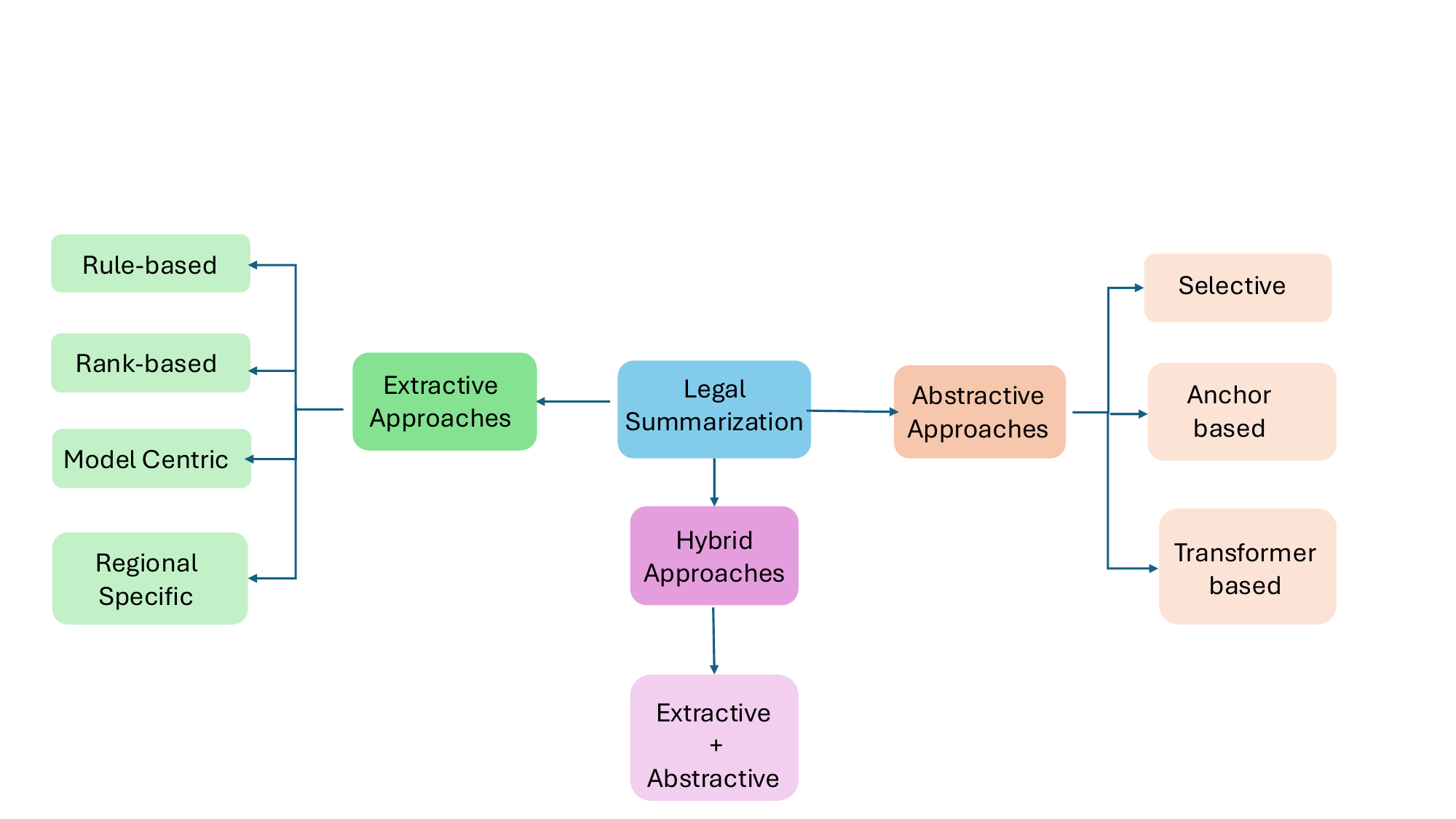} 
    \caption{Taxonomy of legal summarization strategies categorized into three main approaches, each with more specific sub-categories.}
    \label{fig:taxonomy}
\end{figure}

\subsection{Abstractive legal summarization}
Abstractive legal summarization aims to condense extensive legal documents into concise, coherent summaries that capture the essence of the original texts without merely copying them. This task is particularly challenging due to the complexity and technical nature of legal language, which demands a high level of understanding and accuracy in the summarization process. In this section, we will take a broader look at various strategies for abstractive legal summarization.

\paragraph{Selective approaches} These approaches focus on identifying and abstracting the most relevant information from source documents. For instance,~\cite{DBLP:journals/eswa/JainBB24} begins by generating k extractive summaries from the original dataset and then creates new training samples by aligning these extractive summaries with ground truth summaries. Subsequently, a pre-trained BART model is fine-tuned to produce high-quality abstractive summaries. Another approach~\cite{DBLP:conf/ijcnlp/ZhongL23} employs a multi-stage process: in Stage 1, a sentence structure classifier is trained on manually annotated opinion-summary pairs; in Stage 2, the classifier predicts silver labels for unannotated summaries, with special marker tokens guiding the model to adhere to a specific structure. These predicted labels are used to fine-tune an LED model, which is then applied for inference in Stage 3. \citet{DBLP:journals/ail/TranNTS20} segmented documents into three parts, generating summaries for each segment using an extractive summarization model (BSLT) that combines a BERTSUM-based Lawformer encoder with a Transformer architecture to capture document-level features. This approach is further extended with an abstractive summarization model (LPGN), which integrates a pointer-generator network (PGN) with a Lawformer encoder to enhance the accuracy of case descriptions, resulting in higher-quality summaries. Another strategy~\cite{DBLP:conf/acl-nllp/LukoseDJ22} leverages weak supervision to label important sentences based on specific criteria. A BERT-based classifier is subsequently trained on these labeled sentences, followed by a sequence-to-sequence model to generate abstractive summaries. Similar approaches have been utilized in other studies, albeit with slight variations. For instance,~\citet{DBLP:conf/acl/BajajDKKUWBDDM21} employs the GPT-2 model to select relevant sentences, which are then used to create an abstractive summary. In contrast,~\citet{DBLP:conf/icail/XuA23} first classifies text segments as argumentative or non-argumentative, passing only the argumentative segments for summarization.

\paragraph{Anchor-based approaches} Anchor-based approaches leverage specific keywords or topics as "anchors" to guide the summarization process, ensuring that the summarized content remains rooted in the main points of the text. These techniques have become increasingly important in generating abstractive legal summaries. For instance, one study employs special tokens to delineate keywords, facilitating the summarization process~\cite{10.1145/3594536.3595120}. Another approach enhances this method by integrating a transformer model, dividing the process into two distinct stages: an initial stage that extracts keywords and key sentences using a combination of BERT and LSTM technologies, followed by a second stage that generates abstractive summaries utilizing UNILM with attention mechanisms~\cite{huang2023high}. Similarly, topics have also been employed as anchors to improve summarization. For example, a study uses two encoders and one decoder with topic words to enhance summarization based on the Point-Generator Network (PTGEN)~\cite{DBLP:journals/mlc/HuangYGYX20}. Another approach introduces the Element Graph concept to capture topic information, which is then combined with BERT in a dual-encoder framework to generate abstractive summaries~\cite{DBLP:journals/ijon/HuangYGXX21}. These studies highlight the potential of anchor-based methods to effectively bridge the gap between extraction and abstraction in legal summarization tasks.

\paragraph{Transformer-based approaches} Transformer-based approaches, particularly those utilizing the encoder-decoder architecture, are widely employed for abstractive summarization, including in the legal domain. Models such as BART~\cite{DBLP:conf/acl/LewisLGGMLSZ20}, Pegasus~\cite{DBLP:conf/icml/ZhangZSL20}, and T5~\cite{DBLP:journals/jmlr/RaffelSRLNMZLL20} have been effectively applied to generate abstractive legal summaries. For instance, one study trained an abstractive legal summarization model using BART and subsequently integrated it with a legal judgment prediction (LJP) model. A custom loss function was employed to compare embeddings generated by the LJP model based on the summary and the original text~\cite{DBLP:conf/icccnt/SinghalSYP23}. Another study utilized Conditional Random Fields (CRF) and BiLSTM to extract and label 13 rhetorical roles, representing the function of each sentence, which were then used in an ensemble summarization framework combining BART and legal Pegasus to produce abstractive summaries~\cite{muhammed2024impact}. Additionally, T5 was fine-tuned on a custom dataset for abstractive summarization of legal documents, demonstrating its effectiveness in this domain~\cite{DBLP:conf/cit/PrabhakarGP22}.

Legal documents are typically lengthy, posing unique challenges for abstractive summarization. Various approaches have been developed to address these issues. One method involves dividing documents into sections, processing each section through a transformer model to generate candidate summaries, and evaluating these summaries using BERT to select the highest-scoring summary~\cite{DBLP:journals/ail/FeijoM23}. Another approach utilizes a supervised variant of BERTopic to create seven clusters of contracts, followed by employing Legal Pegasus, fine-tuned for the legal domain, to generate summaries for each cluster~\cite{harikrishnan2024topic}. The Longformer~\cite{DBLP:journals/corr/abs-2004-05150} model has also gained attraction for handling long documents which has been applied in legal domain too. In one study, a pretrained Longformer Encoder-Decoder model was fine-tuned on a specific legal dataset for summarization~\cite{sarwar2022text}. Another study used the LED model to generate candidate summaries and employed a scoring function to evaluate argumentative alignment with the input document, selecting the best candidate~\cite{elaraby-etal-2023-towards}. Additionally, a cost-effective pretraining strategy for the Longformer was proposed, utilizing a Replaced Token Detection (RTD) task on legal texts from the "Pile of Law" corpus. This pretrained model was then applied to domain-specific summarization tasks in legal and medical domains~\cite{DBLP:conf/sustainlp/NiklausG23}.

\subsection{Hybrid legal summarization} 
Hybrid approaches to legal summarization combine both extractive and abstractive architectures. The primary concept is similar to selective methods used in abstractive summarization. In this approach, extractive models are trained or fine-tuned for improved extractive summaries, which are then linked to create more effective abstractive summaries. Additionally, some architectural modifications have been implemented for both extractive and abstractive methods to enhance performance.
For example,~\citet{DBLP:journals/ail/MoroPRI24} proposed a transfer learning approach that uses extractive techniques to select sentences, which are subsequently passed through GPT-2 for abstractive summarization, addressing the lack of labeled legal datasets. Another method employed domain-specific training and fine-tuning of Legal Pegasus, combined with chunking long documents and summarizing them using Longformer, followed by an extractive-to-abstractive pipeline~\cite{DBLP:conf/ijcnlp/ShuklaBPMGGG22}. Voting-based ensemble methods have also been introduced, including ranked list-based ensembles (using Borda count and Reciprocal Ranking) and graph-based ensembles, where a graph of sentence similarity identifies key sentences for summarization~\cite{DBLP:journals/ail/DeroyG024}. Feature selection methods that integrate generic text features with legal domain-specific features have further advanced this hybrid approach~\cite{saravade2023improving}. Additionally, models such as BSLT (extractive) and LPGN (abstractive), based on Lawformer, have been designed to address the challenges of long legal documents~\cite{dan2023enhancing}.

Several studies have explored modifications to extractive summarization processes only in hybrid approaches by incorporating rules and advanced machine learning techniques. One approach involves normalizing legal texts using dictionaries for abbreviations and structured elements like articles and sections. The normalized text is then processed using BART for extractive summarization and PEGASUS for abstractive summarization, with extractive BART achieving the best results~\cite{ghosh2022indian}. Another study employs machine learning to label sentences as important or not, followed by LSTMs to summarize these sentences, which are compared against PEGASUS summaries~\cite{ghimire2023too}.

A hybrid approach uses RoBERTa to convert documents into vectorized sentences, which are passed to an extractive model based on Dilated Gated CNN to extract relevant content. The extracted sentences are then fed into T5-PEGASUS for abstractive summarization~\cite{DBLP:conf/ictai/QinL23}. In a different methodology, Ripple Down Rules are applied to classify sentences into 13 rhetorical roles using algorithms such as decision trees, Naive Bayes, SVM, Conditional Random Fields, and BiLSTM~\cite{takale2022legal}. Lastly, an ontology-based approach extracts semantic knowledge from Chinese legal documents, summarizes the content into knowledge blocks, computes similarities between these blocks, and categorizes the documents accordingly~\cite{DBLP:journals/access/MaZM18}. These methods demonstrate the integration of extractive and abstractive approaches improves the quality and relevance of summaries in legal document processing.

\section{Methods for legal summarization}\label{method}
\label{sec:methods}

From the methodological perspective, legal text summarization research works of the last seven years include techniques that have been proposed over a long period of time. The earliest methods we identified are the machine learning or graph-based classifiers proposed invented in the '90s or early '00s. LLM-based techniques proposed in the last years stand on the opposite side.

\paragraph{Tuning large language models} Some papers that utilize LLMs usually tune them with legal corpora in English or some other language. For example, in \cite{10.1145/3594536.3595177} they introduce a legal holding extraction method based on Italian-LEGAL-BERT, a BERT derivative tuned with legal documents in Italian. Furthermore, in \cite{kalamkar-etal-2022-corpus} they create a corpus of legal documents enriched with rhetorical roles. They create a baseline using Legal Pegasus, a fine-tuned version of Pegasus \cite{DBLP:conf/icml/ZhangZSL20}. Tuned Legal Pegasus was also used in \cite{harikrishnan2024topic} to generate summaries of legal documents pre-clustered in seven categories. InLegalBERT and InCaseLawBERT were developed recently by re-training Legal-BERT \cite{paul2023pretrainedlanguagemodelslegal}. They were used in \cite{purnima-etal-2023-citation} to evaluate the role of citing judgments in creating extractive summaries. Several LLMs like LegalBERT, DistilBERT and RoBERTa were also used in \cite{10544203} after being fine-tuned with legal summaries. The authors check if predictions based on summaries are as effective as predictions based on whole legal documents. Finally, T5 fine-tuned with a newly created corpus of 350 judgments is used in \cite{DBLP:conf/cit/PrabhakarGP22} to generate abstractive summaries.

\paragraph{Transformers combined with LSTMs or CNNs} There are other studies that combine Transformer encoders or decoders with LSTM or CNN structures to form specific network architecture. In \cite{huang2023high} for example, they propose a two-stage summarization approach. First, they use a BERT+LSTM model to annotate sentences for indicating which of them should be part of summary. Next, they input the extracted key sentences to an abstractive model based on the unified pre-training language model \cite{10.5555/3454287.3455457} to get the final summary. Authors of \cite{10.5555/3454287.3455457} fuse topic vectors into an LSTM for improving its ability to extract legal text features. Moreover, in \cite{10.1145/3322640.3326728} they adopt a CNN as classifier for iteratively predict summary sentences from legal case documents. They later select a subset of the sentences for the summary using maximal margin relevance from the summarization template and evaluate with ROUGE metrics.

\paragraph{Weak supervision for labeling sentences} Some studies adopt weakly-supervised learning strategies for labeling legal text sentences as important or not.
In \cite{DBLP:journals/corr/abs-2110-01188} for example, they create a dataset and use a weakly-supervised model to automatically label legal text sentences as summary worthy or not. They also experiment with a 2-layer bidirectional LSTM to produce document summaries. Similarly, in \cite{ANAND20222141} they implement a form of weak supervision by automatically labeling the content sentences as important or not using their similarity with headnote sentences their corpus contains. In the next step, they generate the summary documents by using either a feed-forward neural network or an LSTM.

\paragraph{Data preprocessing and augmentation} There are also studies that make some preprocessing of the documents for extracting features, for scoring sentences, or for data augmentation with additional information. In \cite{10.1145/3594536.3595120} for example, they introduce keywords into the summarization models to help them locate and capture key information from long legal texts. Also, in \cite{huang2023high} the authors follow a two stages approach by first selecting key sentences from the legal judgments. In the second stage, they extract keywords related to technical terms in legal texts and introduce them to the summary-generation model. In \cite{polsley-etal-2016-casesummarizer}, they score the sentence relevance using TF-IDF and revise the initial scores based on entities, dates, and proximity of the sentence to the section headings. Finally, in \cite{Andrade2023BB25HLegalSumLB} they first utilize BERT for clustering candidate sentences and then use BM25 method for ranking the candidates and selecting the best ones for the summary.

\paragraph{Benchmarking with multiple methods} A high number of papers perform benchmarks or comparisons using supervised learning. This usually happens in cases when newly created data collections are proposed. In \cite{sharma2023comprehensive} for example, several supervised learning, graph-based or LLM-based techniques are used to provide baselines for a new dataset they propose. Similarly, LexRank is compared with BERT in \cite{info14040250}. There are also studies like \cite{9826728} that use different techniques and comparing their results not just on legal text summarization, but also on other tasks like document classification.

\paragraph{Less frequent methods} Some studies utilize methods which are rarely used for legal text summarization. One example is \cite{10.1145/3462757.3466092} where they make use of ILP (Integer Linear Programming. Another example is \cite{schraagen-etal-2022-abstractive} where reinforcement learning is used. Finally, in \cite{DBLP:journals/ail/DeroyG024} they get predictions from different predictors and aggregate by using voting schemes.

\paragraph{Papers without any method} It is important to note that there are also some works which do not propose or utilize any method at all. For example, in \cite{10.1145/3594536.3595150} the authors involve legal experts to investigate the validity of ROUGE scores in the automatic summarization of legal texts. Similarly, in \cite{Deroy2021AnAS} they analyze expert-generated summaries by comparing them with respective algorithmic summaries, focusing on the important sentences of legal documents that are missed. Similarly, in \cite{Xu_2023} they propose an evaluation framework based on question-answer pairs generated by GPT-4 which cover the main points of the referenced summary. There are also studies like \cite{10.1145/3635059.3635090} where they describe the steps they followed to create a data set of privacy policies in Greek.

\section{Evalaution metrics for legal summarization}\label{metric}
Evaluation is crucial for tracking progress in the field of legal summarization. Evaluation metrics assess the performance of current methods, providing developers with insights on areas for improvement and focus. While automated evaluation metrics are the most popular due to their efficiency in replacing time-consuming human evaluations, human evaluation remains essential for making fair judgments in the legal domain too. In this section, we will discuss the various automated metrics and the human evaluation dimensions we observed during our study.

\subsection{Automated evaluation metrics}
Many automated evaluation metrics for legal summarization have been adapted from general summarization research. While metrics commonly used in summarization are also applied to legal contexts, we have observed key characteristics in the metrics utilized in this field. For instance, some metrics emphasize lexical overlap, while others prioritize semantic similarity or the quality of rankings.

\paragraph{Lexical-overlap based metrics} \textit{ROUGE}~\cite{lin-2004-rouge} is the most popular lexical overlap-based metric for summarization. It measures n-gram overlap between the summary and the reference text and is commonly used in summarization tasks. We have observed that approximately 95\% of method-based papers have utilized the \textit{ROUGE} metric to report the performance of legal summarization methods. In addition to \textit{ROUGE}, other lexical overlap-based metrics such as \textit{BLEU}~\cite{DBLP:conf/acl/PapineniRWZ02} and \textit{METEOR}~\cite{DBLP:conf/acl/BanerjeeL05} have also been adapted for this purpose in some studies~\cite{deroy2023ready, DBLP:conf/icail/XuA23, muhammed2024impact, jain2021automatic, DBLP:journals/ail/DeroyG024, bannihatti-kumar-etal-2022-towards,DBLP:conf/acl-nllp/LukoseDJ22, DBLP:journals/corr/abs-2407-12848, DBLP:conf/jurix/SalaunTLWLB22, DBLP:conf/jurix/SalaunTLWLB22}. BLEU measures n-gram precision originally for machine translation but can also be applied to summarization. \textit{METEOR}, conversely, combines unigram precision and recall with an emphasis on synonyms and stemming.
 
The metrics mentioned are outdated and often do not account for the complexity of the texts. Lexical overlap-based metrics can unfairly penalize paraphrases, and other semantically equivalent variations, often failing to provide a holistic view of the performance of legal summarization methods. ~\citet{10.1145/3594536.3595150} conducted a thorough analysis of the relationship between ROUGE scores and the coverage of legal content. They found that \textit{ROUGE} does not precisely and exhaustively measure legal content, making its reliability questionable. 

\paragraph{Embedding-based metrics} These metrics utilize semantic representations of text to evaluate quality. Among these, \textit{BERTScore}~\cite{DBLP:conf/iclr/ZhangKWWA20} and \textit{BARTScore}~\cite{DBLP:conf/nips/YuanNL21} have been frequently used in studies of legal summarization~\cite{DBLP:conf/ijcnlp/ZhongL23,datta-etal-2023-mildsum,elaraby-etal-2023-towards, DBLP:journals/corr/abs-2407-12848,DBLP:conf/ijcnlp/ShuklaBPMGGG22, bannihatti-kumar-etal-2022-towards, DBLP:conf/sigir/Malik0FRC24}. \textit{BERTScore} measures semantic similarity by comparing the contextual embeddings of tokens generated by BERT models. In contrast, \textit{BARTScore} assesses the likelihood of a summary based on the source or reference using a generative model like BART. Additionally, Cosine Similarity has been employed to assess the quality of legal summary~\cite{jain2021automatic} by computing the cosine distance between vector representations of the summary and the reference or source text. These metrics focus on semantic similarity rather than exact word overlap, making them more robust to paraphrasing and linguistic variations.

\paragraph{Factuality and consistency metrics} Factual consistency in legal summarization is crucial because errors can lead to misinterpretations or unjust outcomes in high-stakes situations, such as contracts or court cases. Maintaining consistency helps build trust in legal AI tools and upholds the ethical standards of the legal profession. Abstractive legal summaries can sometimes produce inaccurate information, which has led to numerous studies supporting the performance of various proposed methods through the use of factual consistency metrics~\cite{DBLP:journals/ail/MoroPRI24, DBLP:conf/ijcnlp/ShuklaBPMGGG22, DBLP:conf/ijcnlp/ZhongL23,deroy2023ready, DBLP:journals/corr/abs-2407-12848}.

These metrics include \textit{FActCC}~\cite{DBLP:conf/emnlp/KryscinskiMXS20}, which assesses factual consistency by verifying information against external sources. \textit{SummaC}~\cite{DBLP:journals/tacl/LabanSBH22} offers a specific approach for evaluating factuality through sentence-level comparisons. Additionally, \textit{NEPrec} assesses the precision of named entities, while \textit{NumPrec} focuses on ensuring numeric accuracy in summaries. Key characteristics of these metrics include evaluating the faithfulness and factual accuracy of generated content, which is critical for abstractive legal summaries that are prone to hallucination.

\paragraph{Ranking metrics} Extractive approaches to legal summarization often involve identifying and ranking relevant information from source documents. Metrics commonly used in ranking, retrieval, and evaluating ordered results can be effectively applied in this context to assess the quality of the ranking mechanisms. For example, some studies have employed ranking and relevance metrics to demonstrate the performance of their proposed methods~\cite{DBLP:conf/emnlp/SanchetiGSR23}. Notably, Mean Average Precision (\textit{MAP})~\cite{DBLP:reference/db/X09xhu}, which evaluates ranking quality based on precision at each relevant document, and Normalized Discounted Cumulative Gain (\textit{NDCG})~\cite{DBLP:journals/tois/JarvelinK02}, which assesses ranking quality by considering both the relevance of items and their positions, have been utilized. These metrics emphasize the importance of relevance and positional significance in evaluating ranked results.

\paragraph{Classification metrics} Extractive legal summarization often involves classifying the worthiness of sentences for inclusion in the summary, determining whether a sentence is important or unimportant, or identifying segments as argumentative or non-argumentative. To evaluate the performance of these classification tasks, metrics such as \textit{F1, Micro F1, and Macro F1} are used as well~\cite{DBLP:conf/emnlp/SanchetiGSR23,10.1145/3462757.3466098,villata2020using,DBLP:journals/clsr/HongC23,ghimire2023too,10544203,harikrishnan2024topic,DBLP:conf/lacci/MedinaOFSROSSFF22}. \textit{Micro F1} aggregates contributions from all classes, making it more sensitive to larger classes, while \textit{Macro F1} calculates the average F1 score across all classes, treating each class equally regardless of size. These metrics provide a comprehensive evaluation of classification performance by balancing precision and recall across different scenarios.

In addition to the discussion above, Appendix~\ref{overview} provides an overview of the various metrics utilized in different works.

\subsection{Human evaluation}

\begin{figure}
    \centering
    \includegraphics[width=0.8\textwidth, , trim=50 50 50 10, clip]{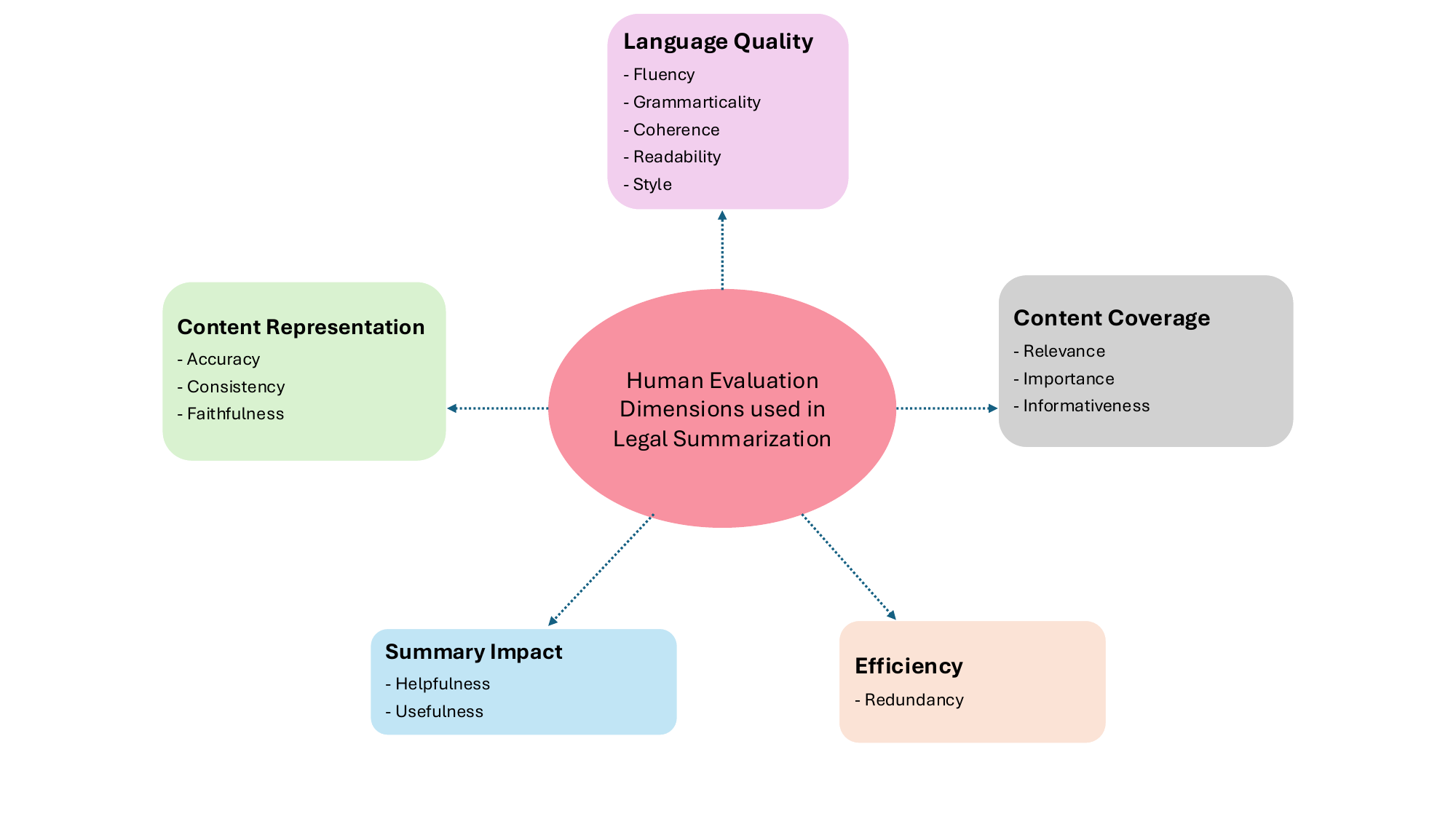} 
    \caption{Overview of key aspects and dimensions in human evaluation for legal summarization}
    \label{fig:humeval}
\end{figure}

As discussed earlier, various evaluation metrics have been utilized for legal summarization. This raises the natural question of which evaluation metric to trust and which one effectively represents progress in legal summarization methods. This issue is not unique to the legal domain; it has also been observed in the broader summarization community~\cite{DBLP:journals/tacl/FabbriKMXSR21}. Unfortunately, there is no single metric that can be relied upon blindly. As a result, researchers frequently turn to additional manual evaluation approaches to further validate their proposed methodologies. While we believe that supporting automated evaluations with human assessments is crucial, only 20\% of studies have backed up their methods with human evaluation. Moreover, only 9\% of these studies have employed experts to conduct this evaluation in the legal domain.

\begin{table}
\begin{small}
\begin{tabular}{lllp{5cm}}
\hline
\textbf{Characteristics} & \textbf{Dimension} & \textbf{Focus} & \textbf{Key Aspect}\\ \hline
\multirow{3}{*}{Content Coverage} & Relevance & Selection of appropriate content & Inclusion of meaningful information\\ 
 & Importance & Prioritization of key content & Summary includes the most significant points from the source \\ 
 & Informativeness & Breadth of relevant information & Amount of meaningful content conveyed \\ \hline
\multirow{2}{*}{Content   Representation} & Faithfulness & Accuracy and truthfulness & Factually consistency of summary with the source text \\  
 & Consistency & Uniformity of information & Summary presents internally consistent information without contradictions \\ \hline
Efiiciency & Redundancy & Avoidance of repetition & Ensures conciseness and efficiency by presenting unique information\\ \hline
\multirow{5}{*}{Language Quality} & Fluency & Smoothness of language & Grammatically correct language is used in summary\\ 
 & Grammaticality & Correct use of grammar & Summary adheres to standard grammar rules\\ 
 & Coherence & Logical connection of ideas & The points in the summary are logically organized and flow naturally\\ 
 & Readability & Ease of understanding & Measures how easy it is for readers to comprehend the summary\\ 
 & Style & Appropriateness of tone & Summary's tone, formality, and language style are suitable for its purpose\\ \hline
Summary Impact & Usefulness & General utility in context & Overall practical value of the summary for a specific user or scenario \\ \hline
\end{tabular}
\end{small}
\caption{Overview of Human Evaluation Dimensions, Focus, and Key Aspects in Legal Summarization}
\label{tab:humeval}
\end{table}
 
We have identified several key characteristics for human evaluation that serve as a meta-evaluation framework in legal summarization, which includes: \textit{\textbf{content representation, efficiency, language quality, content coverage, and summary impact}}. Each characteristic comprises specific dimensions, as shown in Figure~\ref{fig:humeval}, which are essential for evaluating the quality of a summary. For instance, \textit{\textbf{relevance}} ensures the inclusion of information that aligns with the main topics of source document, while \textit{\textbf{importance}} emphasizes whether the presented sentence in model summary was important for a goal-oriented reader even if it was not in the human summary. \textit{\textbf{Informativeness}} evaluates the breadth and depth of meaningful content conveyed, and \textit{\textbf{faithfulness}} ensures the summary remains factually consistent with the source text, avoiding hallucinations or misrepresentations. \textit{\textbf{Accuracy}} verifies the correctness of all details and claims, and \textit{\textbf{consistency}} ensures the absence of internal contradictions. \textit{\textbf{Redundancy}} focuses on eliminating unnecessary repetition, promoting efficiency and conciseness. \textit{\textbf{Fluency}} assesses whether the language used is natural and grammatically sound, contributing to readability, while \textit{\textbf{grammaticality}} ensures adherence to proper language rules. \textit{\textbf{Coherence}} examines the logical flow and organization of ideas, and \textit{\textbf{readability}} considers the ease of comprehension based on structure and vocabulary. \textit{\textbf{Style}} evaluates whether the tone and formality align with the intended purpose or audience, and \textit{\textbf{usefulness}} measures how effectively the summary supports achieving a specific task or goal. Table~\ref{tab:humeval} offers a detailed overview of these dimensions.

The common approach involves asking multiple human annotators or legal experts to evaluate a small sample of model-generated summaries using a Likert scale (ranging from 1 to 5 or 1 to 3). This evaluation focuses on specific dimensions to assess the quality of the legal summaries~\cite{liu2024low,sharma2023comprehensive,10.1145/3594536.3595177,info14040250,DBLP:conf/ijcnlp/ZhongL23,DBLP:conf/emnlp/SanchetiGSR23,Andrade2023BB25HLegalSumLB,schraagen-etal-2022-abstractive,DBLP:conf/emnlp/KurisinkelC22,DBLP:conf/jurix/SalaunTLWLB22,DBLP:journals/corr/abs-2407-12848,DBLP:journals/ail/FeijoM23,polsley-etal-2016-casesummarizer,DBLP:conf/acl-nllp/ZhongL22,DBLP:conf/sustainlp/NiklausG23,DBLP:conf/ecir/BhattacharyaHRP19,DBLP:journals/ijon/HuangYGXX21,takale2022legal,10.1145/3322640.3326728}. However, the meta-evaluation samples in the study did not exceed 50, which we believe is quite small. This limitation is understandable, as obtaining expert annotations is both time-consuming and costly, especially for specific legal documents in a particular language. There needs to be a community effort to create benchmark meta-evaluation datasets for various regional legal documents.

\subsection{Research efforts on metrics in legal summarization}
Legal summarization is a complex task due to the presence of technical jargon and the intricacies of multiple sources in legal documents. It is crucial to ensure factual consistency in legal summarization as well, as inaccuracies can lead to severe legal consequences. Additionally, preserving domain-specific nuances is essential, since even subtle changes in wording or context can alter the meaning of legal statements. There is a significant need for evaluation metrics specifically designed to address the unique challenges of legal summarization. However, there have only been a limited number of research efforts focused on developing evaluation metrics dedicated solely to this area.

Recently,~\citet{Xu_2023} proposed a question-answering approach to evaluate legal summaries. In this method, GPT-4~\cite{DBLP:journals/corr/abs-2303-08774} generates a set of question-answer pairs that encompass the main points and information from the reference summary. GPT-4 is then used to provide answers based on the generated questions related to the summary. In contrast,~\citet{DBLP:conf/lrec/MullickNKPRK22} introduced an intent-based evaluation metric, demonstrating its effectiveness in assessing legal documents. On the other hand,~\citet{elaraby2024adding} focuses on a human evaluation strategy for legal summarization. This paper explores the concept of argument coverage and conducts a human evaluation study where argument roles are treated as atomic units. Additionally,~\citet{DBLP:conf/argmining/YamadaTT17} concentrates on human annotation strategies, specifically for Japanese legal documents. This work proposes an annotation method for Japanese civil judgment documents aimed at creating flexible summaries.

\section{Challenges and future directions}\label{future}

In this section, we will discuss the challenges we observed during our study, and from these observations, we will provide some ideas for possible future directions.

\paragraph{User-specific ground truth summary} It is crucial to consider who the end user of the legal summary is, as summarization needs can vary significantly depending on the user. For example, a judge may be more interested in judicial decisions, while a lawyer would focus on the factual summary of legal documents. Conversely, if the summary is intended for the general public, a plain language approach is essential, with minimal legal jargon. So far, we have observed that legal summaries are primarily written by legal experts, with little emphasis on user-specific legal summaries.

\paragraph{Multi-reference summary dataset} So far, we have observed that all legal summarization datasets include only one reference or ground truth summary. Human summarizers can be biased and may focus on only specific parts of a document when creating a summary. However, a set of documents can have multiple distinct and equally valid summaries. A stable consensus summary can be achieved if a large number of references are gathered. To date, we have not noted any efforts to create a multi-reference summarization dataset in the legal domain.

\paragraph{Domain-specific embeddings in legal summarization} Legal summarization, which is a downstream natural language processing (NLP) task, heavily relies on the accurate representation of legal terms and documents. These legal texts often contain specialized jargon, complex syntax, and domain-specific semantics. However, there has been limited progress in developing embeddings specifically optimized for legal texts, such as Law2Vec~\cite{DBLP:journals/ail/ChalkidisK19}. Existing general-purpose embeddings may not effectively capture the nuanced relationships between legal terms, resulting in performance degradation for legal summarization. To address this issue, there is a need for dedicated efforts to create domain-specific embeddings tailored to legal documents. This can be accomplished through a collaborative, community-driven approach that includes collecting high-quality legal corpora, training specialized embedding models, and benchmarking these models on relevant legal tasks to ensure their effectiveness.

\paragraph{Challenges with lengthy legal documents} Legal documents are inherently lengthy and complex, posing significant challenges for summarization. While researchers have recently begun addressing these challenges, simply applying generic summarization techniques without analyzing their limitations within the legal domain can hinder progress. Current neural summarization models often lack transparency, making it difficult to understand the strategies they employ and assess their effectiveness. Common approaches, such as extractive summarization through document truncation or direct application of abstractive techniques, may fail to capture the full breadth of critical information dispersed throughout legal documents. Furthermore, legal texts often contain dense, interdependent sections where key details may appear in non-contiguous parts of the document. Thus, it is essential to ensure that the generated summaries comprehensively capture and accurately represent the crucial information, while maintaining the logical structure and meaning of the original document. A deeper investigation into model interpretability specific to legal summarization is necessary to address these challenges effectively.

\paragraph{Practicality of legal summarization evaluation metrics} We have observed the use of the ROUGE metric in more than 95\% of the relevant papers. However, the limitations of the ROUGE metric have been widely explored~\cite{DBLP:conf/emnlp/KryscinskiKMXS19,DBLP:conf/acl/MaynezNBM20,DBLP:conf/acl/AkterBS22}. This metric only captures the lexical overlap between reference summaries and model-generated summaries, and it fails to account for the complexities of legal jargon. On the other hand, in the context of abstractive summarization, there should also be a check for hallucinated information. To date, there is no metric that can be blindly trusted. That said, there should be encouragement for the meta-evaluation of model summaries, as well as a more systematic approach to collecting meta-evaluation datasets, similar to SummEval~\cite{DBLP:journals/tacl/FabbriKMXSR21} in the general summarization domain. Although these issues have been recognized to some extent, there has not been a community-wide effort to create benchmarks for evaluation metrics and human evaluation strategies.

\paragraph{Towards multimodal and context-aware legal summarization} A significant limitation in the current research on legal summarization is its narrow focus on summarizing specific types of legal documents, such as court judgments, contracts, and legislative acts. Other important areas, like summarizing legal meetings, dialogues, legal news, legal opinion, court transcripts, and courtroom recordings, have largely been overlooked. Moreover, existing approaches primarily rely on unimodal frameworks that process text alone, without utilizing the potential of multimodal strategies. By integrating multimodal approaches—such as combining textual data with audio or visual content from legal proceedings—it may be possible to create richer and more context-aware summaries. Addressing these gaps, particularly in multilingual and low-resource settings, represents an important opportunity for future research.

\section{Conclusion} \label{conclude}
Through our extensive systematic survey of 123 papers published since the advent of the `transformer era' in 2017, we have identified the common contemporary methods for legal document summarization and highlighted the skewed distribution of countries from which the datasets originate. Our critical analysis revealed gaps in current research, including the lack of multi-reference summaries, the lack of personalization of summaries to possibly diverse target groups, or the almost-exclusive reliance on obsolete summarization metrics such as ROUGE. We also outlined possible future directions including context-aware or multimodal legal summarization.

\section*{Acknowledgements}
This work has been supported by the Research Center Trustworthy Data Science and Security \url{https://rc-trust.ai}, one of the
Research Alliance centers within the \url{https://uaruhr.de}.

\clearpage
\bibliographystyle{ACM-Reference-Format}
\bibliography{main}

\appendix
\section{Appendix}

\subsection{Pretraining corpora}

While not directly related to legal summarization, pretraining corpora are essential for developing models that can effectively process legal text. These corpora establish the foundational linguistic and contextual understanding needed for subsequent fine-tuning on specialized legal datasets.~\autoref{tab:pretraining_corpora} highlights two important pretraining corpora in the legal domain.

\begin{table}[H] \small
    \centering
    \begin{tabular}{l l r r l}
        \toprule
        \textbf{Source} & \textbf{Corpora}       & \textbf{No. of Documents} & \textbf{Total Size in GB} & \textbf{Repository}                 \\
        \midrule
        \multirow{6}{*}{$\text{Legal BERT}$  \cite{chalkidis2020legalbertmuppetsstraightlaw}}
        & EU Legislation        & 61,826                   & 1.9 (16.5\%)             & EURLEX (eur-lex.europa.eu)          \\
        & UK Legislation        & 19,867                   & 1.4 (12.2\%)             & LEGISLATION.GOV.UK                  \\
        & ECJ cases             & 19,867                   & 0.6 (5.2\%)              & EURLEX                              \\
        & ECHR cases            & 12,554                   & 0.5 (4.3\%)              & HUDOC                               \\
        & US court cases        & 164,141                  & 3.2 (27.8\%)             & CASE LAW ACCESS PROJECT             \\
        & US contracts          & 76,366                   & 3.9 (34.0\%)             & SEC-EDGAR                           \\ \hline
       \multirow{4}{*}{$\text{MultiLegalPile}$ \cite{niklaus-etal-2024-multilegalpile}}  & Eurlex Resources &  & 179 (26.0\%) & \href{https://eur-lex.europa.eu/content/legal-notice/legal-notice.html}{eur-lex.europa.eu/content/legal-notice}  \\
     & Native Multi Legal Pile &  & 112 (16.3\%) &  \\
      & Legal mC4 &  & 106 (15.4\%) &  \href{https://huggingface.co/datasets/mc4}{huggingface.co/datasets/mc4}\\
      & Pile of Law &  & 292 (42.4\%) &  \href{https://huggingface.co/datasets/pile-of-law/pile-of-law}{huggingface.co/datasets/pile-of-law}\\
        \bottomrule
    \end{tabular}
    \caption{Pretraining corpora available for legal domain}
    \label{tab:pretraining_corpora}
\end{table}

\subsection{Legal Summarization Overview}\label{overview}

\begin{table*}[h!]
\begin{small}
\begin{tabular}{l l p{2 cm} p{5 cm} p{1.8 cm} p{1.4cm} l}
\hline
\textbf{Work} & \textbf{Year} & \textbf{Method} & \textbf{Description} & \textbf{Dataset} & \textbf{Metric} & \textbf{HumEval} \\ \hline

~\cite{DBLP:journals/ail/JainBB24} & 2024 & DCESumm & 
Identify and rank sentence relevance in a document using Legal BERT, refine scores with deep clustering, and select sorted sentences based on the desired summary length
& BillSum, Forum & ROUGE & No \\ \hline

~\cite{10.1145/3594536.3595177} & 2023	 & Italian-LEGAL-BERT &
Fine-tuned Italian-BERT, Italian-LEGAL-BERT, and Italian-LEGAL-BERT-SC models to predict the most relevant sentences
& ITA CaseHold & ROUGE & Yes\\ \hline

~\cite{Andrade2023BB25HLegalSumLB} & 2023 	& BB25HLegalSum	&
Leverage BERT and BM25 to rank and cluster unique, relevant sentences, aggregate them into candidate summaries, and present the most representative summary by highlighting the most representative sentences.
 & BillSum & ROUGE & Yes\\ \hline

~\cite{purnima2023citation}	 & 2023 & CB-JSumm &
Used InLegalBERT to obtain embeddings for citation and judgment sentences, then applied cosine similarity to select summary sentences. Landmark judgments attract public attention and gather numerous citations, highlighting arguments and precedents that reinforce the cited ruling.
 & IN-Jud-Cit & ROUGE	& No \\ \hline

~\cite{DBLP:conf/jurix/ZinNSSN23} & 2023  &	GPT-3.5	&
Query-based summarization extracts key sentences relevant to predefined queries, which are then processed by GPT-3.5 for information extraction
 & CUAD & F1 & No \\ \hline

 ~\cite{DBLP:conf/iconip/ChenYZ23} & 2023 	&  &
Proposed method comprises a Sentence Encoder, Topic Model, Position Encoder, TS-LSTM Network, Law Article Processor, and Sentence Classifier
 & CAIL & ROUGE & No \\ \hline

~\cite{DBLP:conf/lirai/BauerSGA23}	& 2023	 & Lawformer &
Utilized the Longformer encoder, which was pre-trained on the LegalBART objective using 6 million U.S. court opinions.ion U.S. court opinions
& BillSum & ROUGE	& No \\ \hline

~\cite{rusiya2023implementation} & 2023 & &
Compared BERT, RoBERTa, and XLNet for extractive summarization in the legal domain
&AILA	& ROUGE	& No \\ \hline

~\cite{DBLP:journals/kbs/JainBB23}	& 2023	& Bayesian Optimization Score Fusion &
Current extractive and graph-based methods such as TextRank, LSA, and KLSum have been enhanced with a technique for scoring significant sentences based on linguistic features
& BillSum, GovReport, FIRE, AILA	& ROUGE	& No \\ \hline

~\cite{DBLP:conf/acl-nllp/ZhongL22}	& 2022	& Unsupervised Graph-based Ranking model &

HipoRank employs a reweighting algorithm that considers the history of previously selected sentences to iteratively update sentence importance scores and then select top-k candidates for extractive summary
 
& CanLII &	ROUGE, BERTScore &	Yes \\ \hline

~\cite{10.1145/3462757.3466092} & 2021 & DELSumm & 
An Integer Linear Programming (ILP) objective maximizes summaries by selecting informative sentences, balancing thematic segment representation, and minimizing redundancy, ensuring comprehensive and concise content
& Private & ROUGE & No \\ \hline

~\cite{DBLP:conf/icon-nlp/JainBB21} & 2021 & CAWESumm (Contextual Anonymous Walk Embedding Summarizer) & 
Legal-BERT sentence embeddings and Anonymous Walk Embeddings are concatenated and input into an MLP model to learn binary classification of sentence summary-worthiness during training.
& BillSum & ROUGE & No \\ \hline
 
\end{tabular}
\end{small}
\caption{List of datasets, methods, and metrics related to extractive legal document summarization.}
\label{tab:extractive_overview}
\end{table*}

\begin{table*}[h!]
\begin{small}
\begin{tabular}{l l p{2 cm} p{5 cm} p{1.8 cm} p{1.4cm} l}
\hline
\textbf{Work} & \textbf{Year} & \textbf{Method} & \textbf{Description} & \textbf{Dataset} & \textbf{Metric} & \textbf{HumEval} \\ \hline

~\cite{DBLP:journals/eswa/JainBB24} & 2024 & Extract-then-Assign &
Extractive summaries are generated and matched to ground truth summaries, creating new training samples. BART is then fine-tuned for abstractive summarization using this augmented dataset
& BillSum, FIRE & ROUGE, BERTScore & No \\ \hline

~\cite{harikrishnan2024topic} & 2024 & &
A supervised BERTopic variant clusters contracts into seven groups, followed by Legal Pegasus, fine-tuned for the legal domain, to generate summaries for each document cluster
& Contract Understanding Atticus Dataset (CUAD) & F1 & No \\ \hline

~\cite{huang2023high} & 2023 & &
A two-stage approach: keywords and key sentences are extracted using BERT+LSTM in the first stage, followed by abstractive summary generation with UNILM and attention mechanisms in the second stage
& CAIL, LCRD & ROUGE & No \\ \hline

~\cite{DBLP:conf/ijcnlp/ZhongL23} & 2023 & STRONG (Structure conTRollable OpiNion summary Generation) &
A three-stage process: train a sentence structure classifier on annotated data, predict silver labels for unannotated summaries, and fine-tune the LED model using structure-guided tokens for test set inference
& CanLII & ROUGE, BERTScore & Yes \\ \hline

~\cite{schraagen-etal-2022-abstractive} & 2022 & &
Two methods are proposed: a hybrid reinforcement learning approach combining extractive sentence selection with abstractive rewriting, and a transformer-based summarization method leveraging BART
& RechtspraakNL & ROUGE & Yes \\ \hline

~\cite{DBLP:conf/jurix/SalaunTLWLB22} & 2022 & VanBART, FPTBART&
Court decision summaries are generated based on a question-answer-decision triplet, designed to be intelligible for ordinary citizens without legal expertise
& JusticeBot, CanLII & ROUGE & Yes \\ \hline

~\cite{DBLP:conf/acl/BajajDKKUWBDDM21} & 2021 & BART &
Long documents are compressed by identifying salient sentences, which are then input into BART to generate abstractive summaries
& Amicus data & ROUGE & No \\ \hline

~\cite{DBLP:journals/ijon/HuangYGXX21} & 2021 & ERG-GAT, ERG-GTN, TIG-GAT, TIG-GTN &
A dual-encoder model combines BERT with an Element Graph concept that encodes topic information, enhancing summarization by integrating structured topic representation
LPO-news & ROUGE & Yes \\ \hline

~\cite{DBLP:journals/ail/TranNTS20} & 2020 & &
Documents are segmented into three parts for summary generation. BSLT uses BERTSUM-based Lawformer and Transformer for extraction, while LPGN combines PGN and Lawformer for accurate, high-quality abstractive summaries
& CAIL & ROUGE & No \\ \hline

~\cite{DBLP:journals/mlc/HuangYGYX20} & 2020 & &
Abstractive summarization model with two encoders and one decoder, incorporating topic words to enhance performance, built on the Point-Generator Network (PTGEN) framework
& Private & ROUGE & No \\ \hline

\end{tabular}
\end{small}
\caption{List of datasets, methods, and metrics related to abstractive legal document summarization.}
\label{tab:abstractive_overview}
\end{table*}

\begin{table*}[h!]
\begin{small}
\begin{tabular}{l l p{2 cm} p{5 cm} p{1.8 cm} p{1.4cm} l}
\hline
\textbf{Work} & \textbf{Year} & \textbf{Method} & \textbf{Description} & \textbf{Dataset} & \textbf{Metric} & \textbf{HumEval} \\ \hline

~\cite{DBLP:journals/ail/DeroyG024} & 2024 & Ensemble different summarization algorithms &
Three ensemble methods are proposed: voting-based, ranked-list ensemble (using Borda count and Reciprocal Ranking), and graph-based (leveraging sentence similarity graphs to select identical sentences from connected components) for better summarization
& IN-Ext, IN-abs & ROUGE, METEOR & No \\ \hline

~\cite{dan2023enhancing} & 2023 & BSLT LPGN &
A hybrid legal summarization approach combines the extractive model BSLT and the abstractive model LPGN based on Lawformer to generate the summary
& CAIL2020 & ROUGE & No \\ \hline

~\cite{DBLP:journals/ail/MoroPRI24} & 2023 & Extractive then abstractive &
A transfer learning approach combines extractive and abstractive techniques for legal summarization, addressing limited labeled data by selecting sentences and generating summaries using GPT-2
& Australian Legal Case Report Dataset & ROUGE, FactCC & No \\ \hline

~\cite{ghosh2022indian} & 2023 & Extractive then abstractive &
Legal texts are normalized using dictionaries for abbreviations and article summaries, then processed with BART for extractive summarization and PEGASUS for abstractive summarization
& SCI, IndianKanoon, Manupatra, ILDC & ROUGE & No \\ \hline

~\cite{ghimire2023too} & 2023 & Extractive then abstractive &
Machine learning methods label sentences as important or not, followed by LSTMs for summarization, which are then compared to PEGASUS for performance evaluation
& Opinion of the supreme court of Utah, Idaho, Arizona, New Mexico, Nevada, and Colorado & Recall, F1 & No \\ \hline

~\cite{DBLP:conf/ictai/QinL23} & 2023 & Extractive then abstractive & 
RoBERTa vectorizes document sentences, which are processed by a Dilated Gated CNN extractive model to select relevant sentences. The resulting corpus is then fed into T5 PEGASUS for abstractive summarization
& CAIL & ROUGE & No \\ \hline

~\cite{takale2022legal} & 2022 & Extractive then abstractive &
Ripple Down Rules classify sentences into 13 rhetorical roles using C4.5 decision trees, Naive Bayes, SVM, Conditional Random Fields, and BiLSTM algorithms for improved sentence categorization
& 50 documents of Bombay High Court collected from Legal Search of Manupatra & ROUGE & Yes \\ \hline

~\cite{DBLP:journals/access/MaZM18} & 2018 & Ontology-based &
An ontology-based approach extracts semantic knowledge from Chinese legal documents, summarizes it into knowledge blocks, computes block similarity, and classifies the documents into different categories
& CTA, CDD & Accuracy & No \\ \hline
 
\end{tabular}
\end{small}
\caption{List of datasets, methods, and metrics related to hybrid legal document summarization.}
\label{tab:hybrid_overview}
\end{table*}

\end{document}